\documentclass[10pt,twocolumn,letterpaper]{article}

\usepackage{cvpr}              % To produce the CAMERA-READY version

\usepackage{esvect}
\usepackage{multirow}
\usepackage{cuted}
\usepackage{tabularx}
\usepackage{gensymb}
\usepackage{adjustbox}
\usepackage{array}
\usepackage{booktabs}
\usepackage{commath}
\usepackage{subcaption}
\usepackage{color,soul}
\usepackage{amsmath}
\usepackage[dvipsnames]{xcolor}

\DeclareMathOperator*{\argmax}{arg\,max}

\usepackage{listings}
\lstdefinestyle{jsonbw}{
    language=,
    basicstyle=\small\ttfamily,
    breaklines=true,
    frame=lines,
    showstringspaces=false,
    emph={
        "model", "api_call", "payload", "instructions", "input",
        "role", "content", "type", "text", "image_url", "user",
        context_text, b64_img, b64_img1, b64_img2, category,
        object_name, scheme, dataset, min_frame_gap
    },
    emphstyle=\bfseries,
}

\definecolor{cvprblue}{rgb}{0.21,0.49,0.74}
\usepackage[pagebackref,breaklinks,colorlinks,allcolors=cvprblue]{hyperref}

\def\paperID{26} % *** Enter the Paper ID here
\def\confName{CVPR}
\def\confYear{2026}

\title{Not Your Stereo-Typical Estimator: Combining Vision and Language for Volume Perception}

\author{Gautham Vinod*\quad
Bruce Coburn*\quad
Siddeshwar Raghavan*\quad
Fengqing Zhu
\\
Purdue University, West Lafayette, Indiana, U.S.A.\\
\small{* These authors contributed equally to the work} \\
{\tt\small \{gvinod, coburn6, raghav12, zhu0\}@purdue.edu}
}

\begin{document}
\maketitle

\begin{strip}
\begin{minipage}{\textwidth}\centering
\includegraphics[width=1\textwidth]{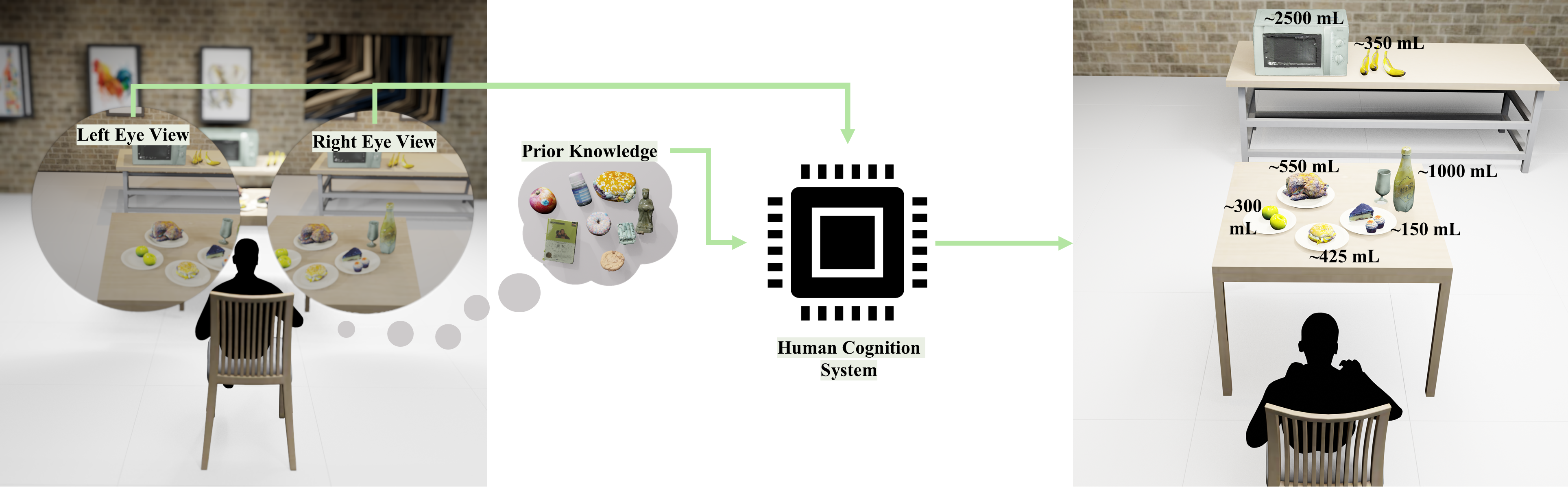}
\captionof{figure}{\textbf{Modeling the Human Cognition System.} Our stereo vision enables us to capture depth information; however, it is our prior knowledge about the shapes and sizes of objects in the scene that enables us to estimate distances and measurements. We leverage the intuition of our vision and cognition system to use stereo images and prior knowledge via natural language to create a multi-modal fusion model for accurate estimation of the sizes of objects.}
\label{fig:intro}
\end{minipage}
\end{strip}

\begin{abstract}
Accurate volume estimation of objects from visual data is a long-standing challenge in computer vision with significant applications in robotics, logistics, and smart health. Existing methods often rely on complex 3D reconstruction pipelines or struggle with the ambiguity inherent in single-view images. To address these limitations, we introduce a new method that fuses implicit 3D cues from stereo vision with explicit prior knowledge from natural language text. Our approach extracts deep features from a stereo image pair and a descriptive text prompt that contains the object's class and an approximate volume, then integrates them using a simple yet effective projection layer into a unified, multi-modal representation for regression. We conduct extensive experiments on public datasets demonstrating that our text-guided approach significantly outperforms vision-only baselines. Our findings show that leveraging even simple textual priors can effectively guide the volume estimation task, paving the way for more context-aware visual measurement systems. \textcolor{cvprblue}{Code: \href{https://gitlab.com/viper-purdue/stereo-typical-estimator}{https://gitlab.com/viper-purdue/stereo-typical-estimator}}.
\end{abstract}
    
\section{Introduction}
\label{sec:intro}

The ability of humans to perceive the world in stereo—relying on the binocular view from the left and right eye—provides an innate and powerful understanding of depth, shape, and size. This stereo vision capability allows us to effortlessly estimate the volume of an object, a crucial skill for understanding and navigating our environment. While monocular cameras allow us to capture the world, the projection of a 3D scene onto a 2D image plane results in the loss of critical depth information, making direct volume estimation a complex and ill-posed problem. Precise volume information is essential across diverse domains, including Medical Imaging and Healthcare~\cite{medical_image_processing, liu2005system}, Food Volume and Nutritional Analysis~\cite{Dehais2017TwoViewFoodReconstruction, he2024MetafoodChallenge}, E-commerce~\cite{e-commerce3d}, Robotics~\cite{robotics_3d}, and Autonomous Systems~\cite{autonomous_3d}.

To address this challenge, traditional methods often rely on structured light or dedicated depth sensors, while modern multi-view approaches such as Neural Radiance Fields (NeRF)~\cite{mildenhall2020nerf} and Gaussian Splatting~\cite{kerbl20233d} attempt to create highly accurate 3D representations from which volume can be derived. Although powerful, these methods often require multiple images or specialized hardware. In the monocular domain, deep learning models have shown promise by learning complex relationships between a 2D image and its volume; however, they remain inherently limited as they must infer the missing third dimension from ambiguous 2D cues.

Stereo vision offers a compelling alternative by mimicking the human ability to perceive depth through stereo disparity. Historically, the primary obstacle to adopting stereo methods was the complexity of calibrating dual-camera setups. Today, this limitation has been effectively eliminated by the ubiquity of multi-camera sensors in modern smartphones and wearable devices, such as Meta's Aria glasses~\cite{engel2023project}. This shift has produced an abundance of stereo data and large-scale egocentric datasets~\cite{perrett2025hd, lv2024aria, pan2023aria}, presenting a clear opportunity to move beyond monocular limitations and leverage readily available geometric cues for precise volume estimation.

In this paper, we propose a multi-modal volume estimator inspired by a fundamental aspect of human cognition: the use of prior knowledge to inform perception. As illustrated in Figure~\ref{fig:intro}, just as humans combine visual depth perception with mental models of familiar objects—instinctively conjuring the typical scale of an ``apple" even without perfect visual input—our approach augments visual data with semantic knowledge. We supply our model with this capability by combining rich geometric cues from a stereo pair with semantic, class-based priors delivered via natural language, inspired by recent advances in Vision-Language Models (VLMs).

Our proposed method leverages pre-trained image encoders to extract features from a stereo pair. Crucially, our model learns the underlying stereo geometry implicitly, removing the need for explicit camera calibration. Simultaneously, we utilize these features to build a stereo image classifier that generates a text prompt containing the object's class and a class-conditional volume prior. These distinct visual and textual features are fused within a VLM-inspired projection layer, creating a hybrid representation that guides the final volume regression with both ``what it sees'' and ``what it knows.''

To the best of our knowledge, this work is the first to propose a multi-modal approach that combines stereo vision with language priors for volume estimation. Our main contributions are:
\begin{itemize}
\item We propose a framework that fuses rich geometric features from stereo images with semantic priors from natural-language text for guided volume estimation.
\item We introduce a powerful projection layer, inspired by VLMs, to effectively combine stereo and text features into a hybrid representation.
\item We demonstrate through extensive experiments on publicly available datasets that our method significantly outperforms existing monocular, stereo, and multi-view volume estimation benchmarks.
\end{itemize}
\section{Related Works}
\label{sec:related_works}

\begin{figure*}[ht!]
    \centering
    \includegraphics[width=1\textwidth]{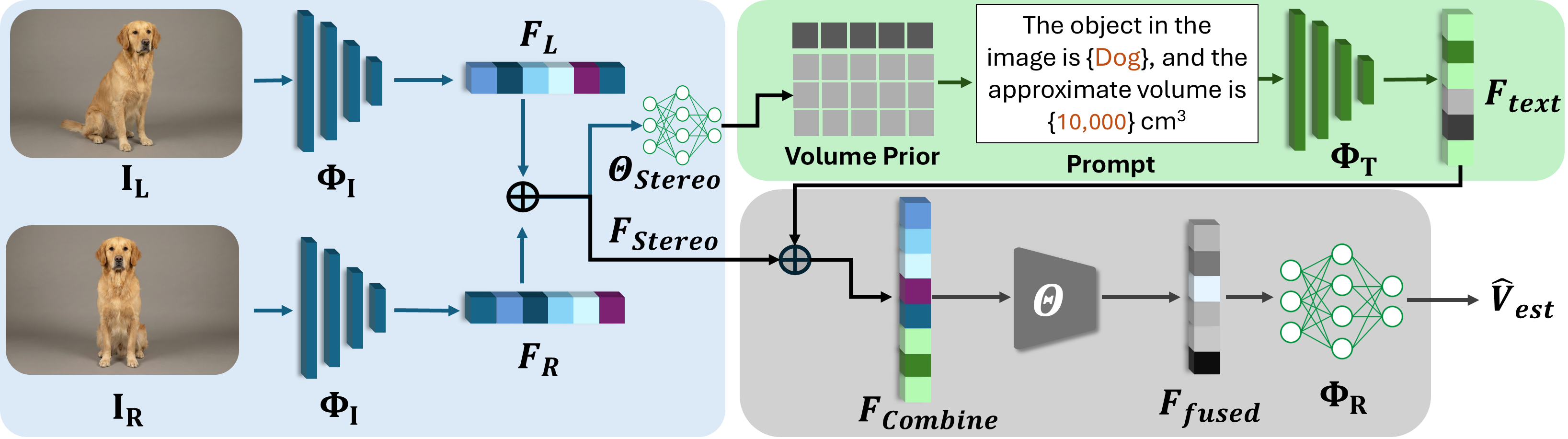}
    \caption{\textbf{Overview of Our Method:} Stereo images are passed through a feature extractor to generate embeddings, which are concatenated and used for classification to obtain the corresponding text prior for each class. The text embedding is generated from the constructed prompt with the text and volume prior. The combined image and text features are projected into a shared space. Finally, a regression network estimates the volume $\hat{V}_{\textit{est}}$.}
    \vspace{-0.5cm}
    \label{fig:overview}
\end{figure*}

\textbf{Volume Estimation Methods.} The task of estimating an object's volume from visual data is a long-standing challenge in computer vision, with applications spanning from medical imaging to robotics~\cite{medical_image_processing, autonomous_3d, vinod2026food}. This task has evolved from template matching~\cite{xu2013Model} or depth sensors~\cite{thames2021nutrition5k} to more data-driven approaches. Multi-view Stereo (MVS)~\cite{schoenberger2016mvs, schoenberger2016sfm} techniques reconstruct dense 3D point clouds from multiple images~\cite{puri2009recognition, Konstantakopoulos2021Stereo, Dehais2017TwoViewFoodReconstruction}, or use implicit 3D representations such as NeRF~\cite{mildenhall2020nerf, kerbl20233d}, from which volume can be computed. These methods require numerous, often calibrated, images of a static scene, limiting their practicality for casual capture. Further, these methods face the issue of losing scale during reconstruction as the generated 3D model exists in a normalized space. 
% Depth-based methods offer a more direct route by using data from RGB-D cameras or LiDAR to create a 3D representation~\cite{lo2019depth, Shao2023VoxelReconstruction}. However, their primary limitation is the dependence on specialized and often expensive sensor hardware.

% Deep learning methods for volume estimation have been developed to utilize more accessible inputs. Many approaches train regression networks using single RGB or RGB-D images~\cite{vinod2022image, thames2021nutrition5k, han2023dpf, Kwan2025Nutrition}. While effective, models using only RGB data are fundamentally ill-posed, as they must infer the missing third dimension from 2D cues, making them sensitive to different viewpoints and object variations. Models using RGB-D data perform better but retain the hardware dependency of depth sensors or must rely on depth estimation networks~\cite{han2023dpf, chen2024metafood3d}. Recently, large foundation models such as GPT-5~\cite{OpenAI2025GPT5SystemCard} with vision have demonstrated remarkable zero-shot reasoning capabilities. However, when prompted for quantitative volume estimation, their performance is often unreliable, as they lack the explicit 3D geometric grounding required for precise metric prediction, a limitation we confirm in our experiments.

To overcome the multi-view constraint, monocular deep learning methods infer volume from single RGB or RGB-D images~\cite{vinod2022image, thames2021nutrition5k, han2023dpf, Kwan2025Nutrition, vinod2026size}. While RGB-D approaches perform well, they depend on depth sensors. Purely RGB methods are more accessible but inherently ill-posed, as they must hallucinate depth from ambiguous 2D cues. Even recent multimodal foundation models like GPT-5~\cite{OpenAI2025GPT5SystemCard}, despite their semantic reasoning, lack the explicit geometric grounding necessary for precise quantitative estimation, a limitation we address by reintroducing stereo constraints.

\textbf{Vision-Language Models.} VLMs like CLIP~\cite{radford2021learning} and Flamingo~\cite{alayrac2022flamingo} have revolutionized semantic tasks by aligning image and text representations in a shared space. However, this success has largely been confined to classification, VQA, and captioning~\cite{liu2023visual}. The application of VLM architectures to metric regression tasks remains underexplored. Unlike general-purpose VLMs that prioritize high-level semantics, our approach adapts cross-modal fusion to combine the precise geometric cues of stereo vision with textual priors specifically for volume regression.

\section{Method}
\label{sec:method}

The core of our proposed method is a multi-modal architecture designed to model the human cognitive process of combining visual perception with prior knowledge for reliable estimation. Our framework, illustrated in Figure~\ref{fig:overview}, systematically processes and fuses information from two distinct modalities, the geometric cues from a stereo image pair and the semantic context provided by a natural language prompt. The pipeline is composed of three sequential modules: (1) a Stereo Module to extract rich visual and geometric representations, (2) a Textual Module to encode class-conditional priors, and (3) a Cross-Modal Fusion Module that integrates these diverse information streams to regress the final volume estimate.

\subsection{Stereo Module: Extracting Geometric Representations}

The primary objective of the Stereo Module is to convert the input stereo image pair, $(I_L, I_R)$, into a feature representation that is rich in both semantic content and implicit geometric information. Unlike monocular images, a stereo pair encodes crucial depth information through disparity, which is the difference in the horizontal position of an object in the two images. Our goal is to leverage this information without requiring explicit depth map computation or camera calibration, allowing the network to learn the underlying epipolar geometry directly from the data.

To achieve this, we first employ a powerful, pre-trained vision backbone, such as a Vision Transformer (ViT)~\cite{dosovitskiy2021vit}, to act as our image feature encoder ($\Phi_I$). We use a shared-weight encoder for both the left and right images. The choice of a large, pre-trained model is deliberate, as these models have learned to extract powerful geometric cues from vast datasets, enabling them to embed high-level semantic features (e.g., object parts, textures) that are essential for object recognition. In the case of traditional stereo approaches, poor stereo matching causes massive artifacts and sometimes failure to generate geometric representations. Our shared encoder approach also allows us to mitigate these conditions by relying on a single image if the other image is noisy.
Each image is passed independently through the encoder to produce high-dimensional feature vectors
% $$ F_L = \text{Encoder}(I_L) $$
% $$ F_R = \text{Encoder}(I_R) $$
$$ F_L = \Phi_I(I_L) $$
$$ F_R = \Phi_I(I_R) $$
where $F_L, F_R \in \mathbb{R}^D$ are the feature representations for the left and right images, respectively.

These two feature vectors are then concatenated to form the final stereo representation, $F_\text{stereo}$,
% $$ F_\text{stereo} = \text{concatenate}(F_L, F_R) $$
$$ F_\text{stereo} = [F_L; F_R] $$
This simple concatenation strategy is highly effective as it preserves all the information from both viewpoints and creates a combined feature vector of dimension $2 \times D$. It allows the subsequent fusion module to learn the complex relationships and correspondences between the left and right features, implicitly discovering the geometric cues necessary for depth perception and volume estimation.

To extract the strong prior information that forms one of the core idea of our model, we need to classify the image to understand what the image contains. Here, we leverage the concatenated image features $F_\text{stereo}$ to serve as the feature space for image classification using stereo images. The concatenated features are passed through a classification head $\Theta_\text{Stereo}(\cdot)$ that predicts the class probability distribution, which is used to obtain the object class label $\hat{y}$,

\begin{equation}
    \hat{y} = \argmax_{x} \Theta_\text{Stereo}(F_\text{Stereo})
\end{equation}

We construct a lookup table using the volume prior information categorically from large scale 3D datasets and publicly available information. Then, the predicted class is used to obtain the corresponding class-conditioned volume prior from the lookup table for prompt construction in the textual module.

\subsection{Textual Module: Injecting Semantic Priors}
The intuition behind the Textual Module is to provide the model with ``common sense" knowledge, similar to how a human leverages past experience. For instance, knowing an object is a ``mug" immediately constrains our expectation of its volume to a certain plausible range. This module formalizes this intuition by injecting a class-conditional volume prior into the model via natural language.

First, we establish a data-driven prior by creating a lookup table of the mean volume for each object category present in our training dataset. This provides a reasonable initial guess, or anchor, for the volume based solely on the object's class.

Next, we use this information to construct a structured, descriptive text prompt. For a given object, we use its known class (or the output of a classification model) and its corresponding mean volume from the lookup table to generate a sentence following a specific template:

\begin{quote}
    ``These are stereo image pairs of \{\ \textbf{object class} \}\ whose approximate volume is \{\ \textbf{mean volume} \}\ mL.''
\end{quote}

This prompt-based approach is significantly more powerful than simply using one-hot encoded class labels or a raw volume number. By embedding the information in a natural language sentence, we can leverage the sophisticated understanding of language inherent in modern pre-trained text encoders (e.g., from CLIP~\cite{radford2021learning} or BERT~\cite{devlin2019bert}). The structure of the sentence provides rich context, confirms the input modality (``stereo image pair"), identifies the object, and provides a numerical anchor for its volume.

This prompt is then fed into a pre-trained text encoder ($\Phi_T$), which converts it into a dense feature vector, $F_\text{text}$, that captures the semantic meaning of the prior information.
% $$ F_{\text{text}} = \text{TextEncoder}(\text{prompt}) $$
$$ F_{\text{text}} = \Phi_T(\text{prompt}) $$

The impact of this class conditioned prior injection is shown experimentally in Section~\ref{sec:feature_combination}. 

\subsection{Cross-Modal Fusion and Volume Regression}
At this stage, we have two powerful but separate representations, $F_\text{stereo}$ capturing the ``what it looks like" information and $F_\text{text}$ providing the ``what it should be like" contextual direction. These features exist in different embedding spaces and must be intelligently combined. Our fusion module, inspired by the projection layers in VLMs, is designed to map these distinct features into a shared, unified space where they can interact and collectively inform the final volume prediction.

\begin{figure}[h!]
    \centering
    \includegraphics[width=0.8\linewidth]{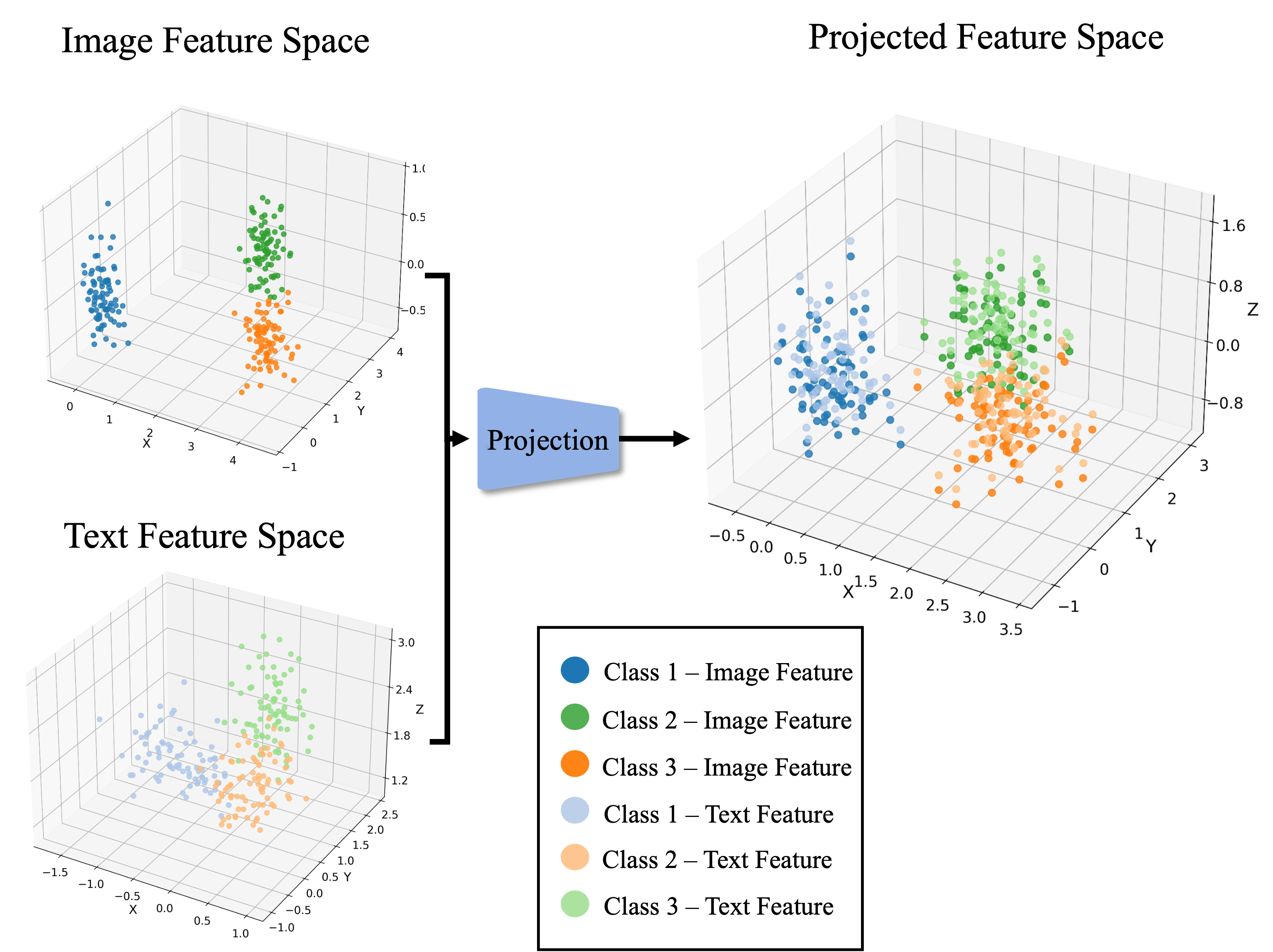}
    \caption{\textbf{Visualization of our multimodal projection -} The image and text features each form three separate clusters corresponding to the same class. Our projection maps these modalities into a unified space, where semantically similar image and text features cluster together.}
    \label{fig:projection}
\end{figure}

The fusion process begins by concatenating the stereo image features and the text features.
 \begin{equation}
    F_{\text{combine}} = [F_{\text{stereo}}; F_{\text{text}}]
\end{equation}

Next, we project the combined visual and textual features into a common latent space of dimension $K$ using a Multi-Layer Perceptron (MLP) $\theta$ as shown in Figure~\ref{fig:projection}. $\theta$ consists of a linear layer with non-linear activation function (ReLU), allowing the model to learn the complex transformations required for alignment. The projection of the fused features creates a unified multi-modal representation.

 \begin{equation}
    F_{\text{fused}} = \theta(F_{\text{combine}})
\end{equation}

%  \begin{equation}
%     P_{\text{stereo}} = \Theta_{\text{I}}(F_{\text{stereo}})
% \end{equation}
% \begin{equation}
%     P_{\text{text}} = \Theta_{\text{T}}(F_{\text{text}})
% \end{equation}

% where the projected features $P_{\text{stereo}}, P_{\text{text}} \in \mathbb{R}^K$. The projection MLP consist of a linear layer with non-linear activation function (ReLU), allowing the model to learn the complex transformations required for alignment.

% Once in the common space, the features are fused using element-wise addition to create a unified, multi-modal representation, $F_{\text{fused}}$:
% \begin{equation}
%     F_{\text{fused}} = P_{\text{stereo}} + P_{\text{text}}
% \end{equation}

% This additive fusion allows the textual prior to directly modulate the visual features, effectively guiding the network's attention to relevant visual cues.

Finally, the fused feature vector $F_{\text{fused}}$ is passed to a \textbf{regression head}, which is another MLP ($\Theta_{\text{R}}$), to predict the final volume $\hat{V}_{\textit{est}}$.
\begin{equation}
    \hat{V}_{\textit{est}} = \Theta_{\text{R}}(F_{\text{fused}})
\end{equation}
The regression head maps the high-dimensional fused representation to a single scalar value representing the estimated volume. The Mean Squared Error (MSE) loss is chosen for the regression and is defined as the MSE between the predicted volume $\hat{V}_{\textit{est}}$ and the ground-truth volume $V_{\textit{gt}}$:
\begin{equation} \label{eq:mse_loss}
    \mathcal{L}_{\text{MSE}} = \frac{1}{N} \sum_{i=1}^{N} (V_{\textit{gt}}^{i} - \hat{V}_{\textit{est}}^{i})^2
\end{equation}
where $N$ is the number of samples in a batch.

The stereo image classifier uses the Cross-Entropy Loss:
\begin{equation}\label{eq:ce_loss}
    \mathcal{L}_\text{CE} = \frac{1}{N}\sum_{i=1}^N \left[ \sum_{c=1}^C y_c \log(p_c) \right]
\end{equation}
where $C$ is the number of classes, $y_c$ is a binary indicator and is 1 when the sample $i$ belongs to class $c$ and $p_c$ is the probability of that sample $i$ belongs to class $c$.

The whole method is trained end-to-end and supervised on the losses in Equations~\ref{eq:mse_loss} and \ref{eq:ce_loss} as:
\begin{equation} \label{eq:loss}
    \mathcal{L} = \lambda \mathcal{L}_{\text{MSE}} + \mu \mathcal{L}_{\text{CE}}
\end{equation}
where $\lambda$ and $\mu$ are the weights for the loss functions.
\section{Experimental Results}
\label{sec:exp_results}

\begin{table*}[ht!]
    \centering
    \resizebox{\textwidth}{!}{
    \renewcommand{\arraystretch}{1.1}
    \setlength{\tabcolsep}{5pt}
    \begin{tabular}{r | c c c c c | c c c c c}
    \toprule
    \multirow{2}{*}{\textbf{Method}} & \multicolumn{5}{c|}{\textbf{MetaFood3D Dataset}} & \multicolumn{5}{c}{\textbf{OmniObject3D Dataset}} \\
     & MAE (mL) $\downarrow$ & MAPE (\%) $\downarrow$ & r $\uparrow$ & $R^2$ $\uparrow$ & $\cos(\theta)$ $\uparrow$ & MAE (mL) $\downarrow$ & MAPE (\%) $\downarrow$ & r $\uparrow$ & $R^2$ $\uparrow$ & $\cos(\theta)$ $\uparrow$ \\
    \midrule
    Baseline          & 164.19 & 526.79          & 0.00          & -0.01         & 0.68          & 140.28          & 444.57          & 0.00          & -0.02         & 0.60 \\
    Category Mean     & 134.86 & 202.88          & 0.46          & 0.07          & 0.72          & 118.47          & 87.46           & 0.63          & 0.35          & 0.76 \\
    RGB Est.~\cite{thames2021nutrition5k}          & 151.98 & 64.65           & 0.57          & -0.37         & 0.80          & \underline{79.53} & 68.34           & \underline{0.81} & \underline{0.46} & \underline{0.85} \\
    Depth Recon.~\cite{fang2016comparison}      & 210.45 & 287.63          & 0.54          & -1.08         & 0.78          & 125.34          & 175.97          & 0.46          & 0.21          & 0.61 \\
    3D Assisted~\cite{Vinod20243DFoodPortion}       & 134.52 & 97.84           & 0.48          & -0.57         & 0.68          & -               & -               & -             & -             & - \\
    GPT-5~\cite{openai_gpt_api}             & 92.38  & 198.83          & 0.65          & 0.39          & 0.82          & 101.21          & 62.33           & 0.61          & 0.25          & 0.71 \\
    GPT-5 (w/context)~\cite{wei2022chain} & \underline{90.59} & \underline{53.86} & \underline{0.66} & \underline{0.39} & \underline{0.82} & 93.70           & \underline{57.50}  & 0.60          & 0.31          & 0.71 \\
    NeRF~\cite{mildenhall2020nerf}              & 114.26 & 142.04          & 0.37          & 0.00          & 0.73          & 135.89          & 237.56          & 0.17          & 0.02          & 0.66 \\
    \midrule
    \textbf{Ours}     & \textbf{47.95} & \textbf{22.54} & \textbf{0.88} & \textbf{0.75} & \textbf{0.94} & \textbf{44.92} & \textbf{29.74} & \textbf{0.91} & \textbf{0.82} & \textbf{0.93} \\
     & (\textbf{-47.1\%}) & (\textbf{-58.2\%}) & (\textbf{+33.3\%}) & (\textbf{+92.3\%}) & (\textbf{+14.6\%}) & (\textbf{-43.5\%}) & (\textbf{-48.3\%}) & (\textbf{+12.3\%}) & (\textbf{+78.3\%}) & (\textbf{+9.4\%}) \\
    \bottomrule
    \end{tabular}}
    \caption{\textbf{Volume Estimation Comparison.} We evaluate performance on the MetaFood3D and OmniObject3D datasets using five metrics: MAE, MAPE, Pearson correlation ($r$), $R^2$, and cosine similarity ($\cos(\theta)$). Our method consistently outperforms all baselines. The next-best methods are \underline{underlined}. The ``Baseline'' refers to predicting the dataset's mean volume for all inputs. The ``Category Mean'' predicts the mean volume of each sample's category. Results from ``3D Assisted'' are omitted for OmniObject3D as it requires a checkerboard pattern absent in that dataset.}
    \vspace{-0.3cm}
    \label{tab:volume_comparison}
\end{table*}

\begin{table}[ht!]
    \centering
    \renewcommand{\arraystretch}{1.1}
    \setlength{\tabcolsep}{5pt}
    \resizebox{\linewidth}{!}{
    \begin{tabular}{r | c c c c}
    \toprule
    \textbf{Stereo Methods} & \textbf{MAE (mL)} $\downarrow$ & \textbf{MAPE (\%)} $\downarrow$ & $\mathbf{r}$ $\uparrow$ & $\boldsymbol{\cos(\theta)}$ $\uparrow$ \\
    \midrule
    Baseline          & 151.85   & 845.69   & 0  & 0.707 \\
     Category Mean     & 134.86 & 202.88          & 0.46         & 0.72  \\
    SIFT Recon.~\cite{Dehais2017TwoViewFoodReconstruction}       & 345.34   & 610.92   & -0.01      & 0.17 \\
    ORB Recon.~\cite{Rublee2011Orb}        & 2718.38  & 870.80   & 0.14      & 0.22 \\
    LightGlue Recon.~\cite{lindenberger2023lightglue}  & 203.85   & 221.13   & 0.18      & 0.46 \\
    NeRF Stereo~\cite{mildenhall2020nerf}       & 104.59    & 132.67   & 0.56   & 0.81 \\
    GPT-5 Stereo~\cite{openai_gpt_api}      & 71.30    & 42.88    & 0.78      & 0.89 \\
    \textbf{Ours}     & \textbf{44.92}  & \textbf{29.74}  & \textbf{0.91} & \textbf{0.93} \\
    \bottomrule
    \end{tabular}}
    \caption{\textbf{Stereo Volume Estimation Comparison.} We evaluate performance on MetaFood3D using four metrics: MAE, MAPE, Pearson correlation ($r$), and cosine similarity ($\cos(\theta)$). Our method consistently outperforms all baselines. The ``Baseline'' refers to predicting the dataset's mean volume for all inputs. The ``Category Mean'' predicts the mean volume of each sample's category.}
    \vspace{-0.7cm}
    \label{tab:stereo_comparison}
\end{table}

% \begin{itemize}
%     \item Datasets: Describe the datasets MetaFood3D and OmniObject3D used for evaluation.
%     \item Implementation Details
%     \item Evaluation Metrics
%     \item Main results - Table 1
%     \item Stereo Reconstruction Results - Table 2
% \end{itemize}

\subsection{Datasets}
Our proposed method is evaluated using accurate ground-truth volume data extracted from the high-quality 3D meshes in publicly available OmniObject3D~\cite{wu2023omniobject3d} and MetaFood3D~\cite{chen2024metafood3d} datasets. 
MetaFood3D~\cite{chen2024metafood3d} contains highly accurate 3D models of food, we use 529 items for training and 102 items in our testing sets. 
OmniObject3D~\cite{wu2023omniobject3d} provides a large number of general purpose 3D models from which we use 2,763 in the training set and 654 in the testing set.
Both datasets contain video frames where the camera moves around the stationary foreground object. We sample 2 non-consecutive frames to simulate a stereo vision setting that includes some disparity between the frames. 

During inference, only a pair of images is used per item in the dataset. During training, we leverage the availability of multiple frames to train the model using different combinations of ``left'' and ``right'' stereo images.

\subsection{Metrics}
To evaluate our model's performance, we rely on standard regression metrics - Mean Absolute Error (MAE), Mean Absolute Percentage Error (MAPE), Pearson Coefficient ($r$), Coefficient of Determination $R^2$, and cosine similarity $\cos(\theta)$ which are defined as follows.

\vspace{-0.4cm}
{\small
\begin{alignat*}{2}
    &\text{MAE} = \frac{1}{N}\sum_{i=1}^N |\hat{\mathbf{V}}_{\textit{est}}^i - \mathbf{V}_{\textit{gt}}^i|, \quad
    &\text{MAPE} = \frac{1}{N}\sum_{i=1}^N \frac{|\hat{\mathbf{V}}_{\textit{est}}^i - \mathbf{V}_{\textit{gt}}^i|}{\mathbf{V}_{\textit{gt}}^i},
\end{alignat*}
}
\vspace{-0.5cm}

{\small
\begin{align*} % Newly formatted equation block
&r = \frac{\sum_i (\hat{\mathbf{V}}_{\textit{est}}^i - \bar{\mathbf{V}}_{\textit{est}})(\mathbf{V}_{\textit{gt}}^i - \bar{\mathbf{V}}_{\textit{gt}})} {\sqrt{\sum_i{(\hat{\mathbf{V}}_{\textit{est}}^i - \bar{\mathbf{V}}_{\textit{est}})}^2{\sum_i {(\mathbf{V}_{\textit{gt}}^i - \bar{\mathbf{V}}_{\textit{gt}})}^2}}},
\end{align*}
\begin{alignat*}{2}
    % &R^2 = r^2, \quad
    & R^{2} &= 1 - \frac{\sum_i \big(\mathbf{V}_{\textit{gt}}^i - \hat{\mathbf{V}}_{\textit{est}}^i)^2}{\sum_i \big(\mathbf{V}_{\textit{gt}}^i - \bar{\mathbf{V}}_{\textit{gt}})^2}, \quad
    &\cos(\theta) = \frac{\vv{\mathbf{V}}_{\textit{est}} \cdot \vv{\mathbf{V}}_{\textit{gt}}}{\norm{\vv{\mathbf{V}}_{\textit{est}}} \norm{\vv{\mathbf{V}}_{\textit{gt}}}}
\end{alignat*}
}
where $N$ is the number of 3D models in the dataset, $\hat{\mathbf{V}}_{\textit{est}}^i$ is the estimated volume, $\mathbf{V}_{\textit{gt}}^i$ is the ground truth volume of the object, $\bar{\mathbf{V}}_{\textit{est}}$ is the mean of the estimated volumes across the dataset, and $\bar{\mathbf{V}}_{\textit{gt}}$ is the mean of the ground-truth 3D volumes across the dataset. $\vv{\mathbf{V}}_\textit{est}$ is the vector of all volume estimates in the dataset and $\vv{\mathbf{V}}_\textit{gt}$ is a vector of all ground-truths for the cosine similarity metrics. 

The units of volume in both datasets are scaled to milliliters (mL). The MAPE is used to determine the best-performing method mainly due to the prevalence of low ground-truth values in the datasets, which causes spikes in the MAPE. The Pearson coefficient ($r$) and the coefficient of determination ($R^2$) assess the correlation between the estimates and ground truth, and cosine similarity measures the alignment of the estimation trend with the ideal $y=x$ line.

\vspace{-0.1cm}
\subsubsection{Implementation Details}
The stereo image encodings $F_L$ and $F_R$ are obtained using a frozen ViT-L/14@336px~\cite{dosovitskiy2021vit} based CLIP~\cite{radford2021learning} image encoder. The text features are extracted using a BERT based sentence transformer~\cite{reimers-2019-sentence-bert}, specifically the MPNet-v2 encoder~\cite{song2020mpnet}. The output image embeddings and text embeddings have a dimensionality of 768. 

The model was trained for 100 epochs with a batch size of 64 using the Adam optimizer ($\beta_1=0.9, \beta_2=0.999$) with a learning rate of $\alpha=0.001$ on a single NVIDIA A40 GPU. We use $\lambda=1$ and $\mu=0.5$ to weight the loss functions in Equation~\ref{eq:loss} that were determined by trial and error.

\subsubsection{Methods of Comparison}
We compare our method against many established baselines for volume estimation as detailed below. The chosen methods vary in the inputs that they require from monocular images, depth maps, and 3D models to multi-view inputs. To ensure a fair comparison, all methods have been trained or fine-tuned on the same training data in each dataset. 
Since reconstruction-based methods typically use size references from the image, the mean scaling factor between the reconstructed volumes and the ground-truth for each category is used to scale the reconstructed models to real-world units and estimate their volume.

\textbf{RGB Estimation}~\cite{thames2021nutrition5k} - A baseline model that directly regresses volume from a single RGB image using a lightweight feature extractor followed by MLP layers.

\textbf{Depth Reconstruction}~\cite{fang2016comparison} - Generates a depth map from the RGB image using a monocular depth estimator, Metric3D~\cite{hu2024metric3d}, then reconstructs a voxel model. Ground-truth volumes are used to compute a dataset-wide average scale factor for conversion to real-world units.

\textbf{3D Assisted Estimation}~\cite{Vinod20243DFoodPortion} - Matches the input image to a known 3D template model using geometric cues and a reference checkerboard for scale. Not applicable to OmniObject3D~\cite{wu2023omniobject3d} due to the lack of reference markers.

\textbf{GPT-5} - Uses the GPT-5 multimodal model~\cite{OpenAI2025GPT5SystemCard} to estimate volume from a single image. The prompt instructs the model to output only the object’s volume (in mL), relying purely on visual input.

\textbf{GPT-5 (w/ context)} - Extends GPT-5 by including textual context in the prompt (\textit{e.g.}, “an image of a banana”) to improve estimation accuracy~\cite{brown2020language, wei2022chain}, leveraging prior knowledge of the object class. More information on the prompts used for the GPT-5 model can be found in the Supplementary Material (Section~\ref{sec:gpt-5}).

\textbf{NeRF}~\cite{mildenhall2020nerf} - A powerful 3D reconstruction tool that uses multiple image inputs from different viewpoints to output a realistic surface reconstruction of the object. 

\begin{figure}
    \centering
    \includegraphics[width=0.8\linewidth]{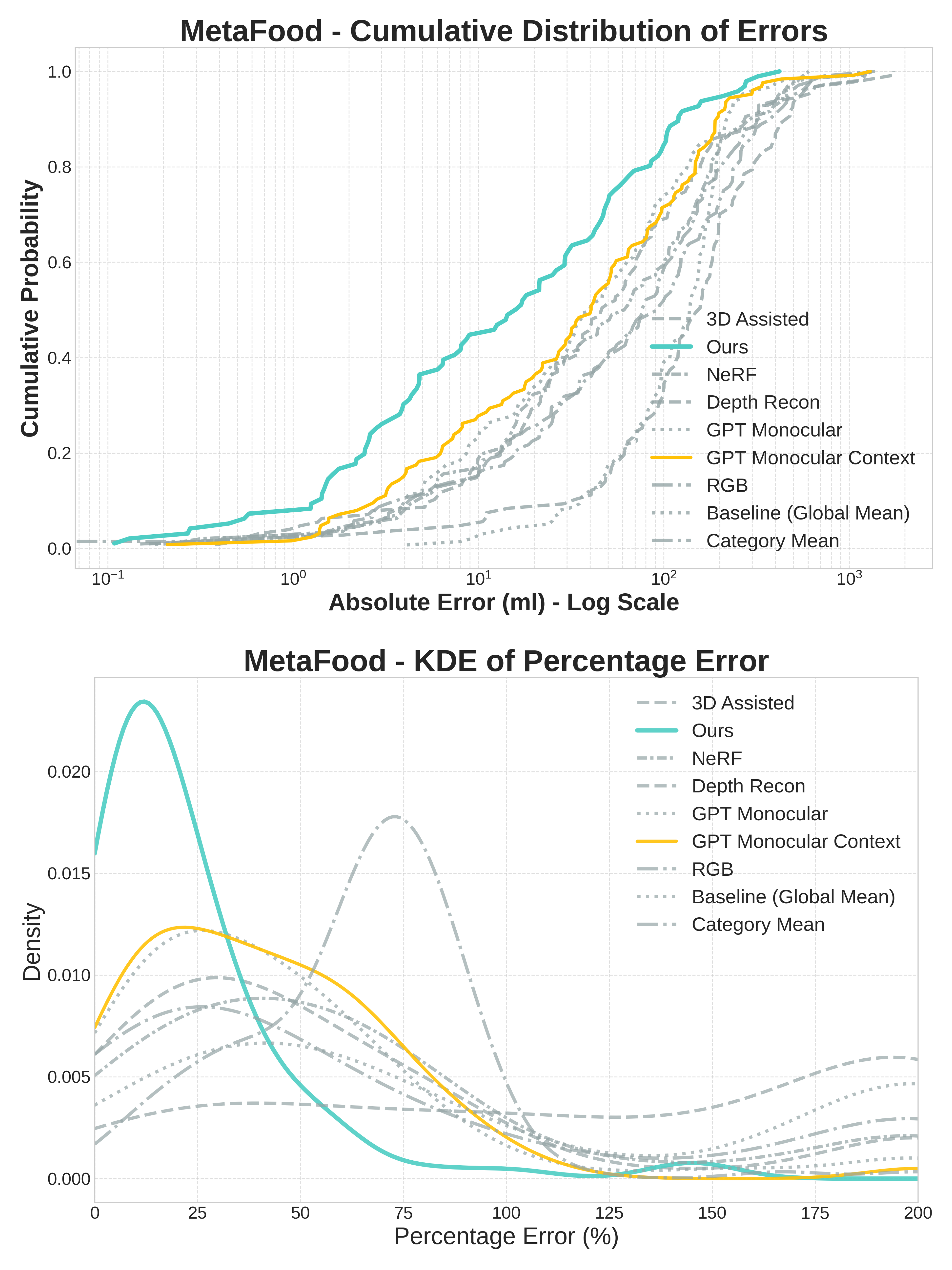}
    \caption{\textbf{Error Distribution of Volume Estimation Methods.} We highlight our method (blue) against the second-best performing method, GPT-5 with context (orange), with regards to MAPE. \textbf{(Top)} The Cumulative Distribution Function (CDF) of absolute errors on a log scale where a shift or curve to the upper left indicates better performance. \textbf{(Bottom)} The Kernel Density Estimation (KDE) of percentage errors where a tall peak near zero with significant delay indicates better performance.}
    \label{fig:big_metafood}
\end{figure}

Table~\ref{tab:volume_comparison} presents a quantitative comparison against a comprehensive suite of baselines, including traditional methods and large vision-language models like GPT-5. Our model achieves remarkable performance, reducing the MAPE by nearly 50\% and leading in all correlation metrics. To ensure a direct comparison against the same input modality, we also benchmark against strong stereo reconstruction techniques using various feature matchers (SIFT~\cite{lowe2004distinctive}, ORB~\cite{Rublee2011Orb}, LightGlue~\cite{lindenberger2023lightglue}) and a two-view NeRF~\cite{mildenhall2020nerf} implementation. As shown in Table~\ref{tab:stereo_comparison}, our method significantly outperforms these specialized stereo baselines as well. An error distribution analysis in Figure~\ref{fig:big_metafood} further highlights our model's performance compared to other methods.

\noindent \textbf{On the Efficacy of Semantic Priors.} On the MetaFood3D dataset in Table~\ref{sec:exp_results},simply predicting the Category Mean yields a massive error of 134.86 mL MAE and 202.88\% MAPE. In stark contrast, our multi-modal framework achieves 47.95 mL MAE and 22.54\% MAPE. We observe similarly drastic error reductions on the OmniObject3D dataset. This massive quantitative gap of more than 88\% reduction in error proves our method does not merely take a shortcut using the text prior. Instead, it actively utilizes the geometric stereo features to perform precise, instance-level volumetric refinement. 
 
Beyond performance on standard benchmarks, we evaluated the model's generalizability through both random-split experiments and a cross-dataset validation. Our method demonstrated strong generalization, notably outperforming even GPT-5 in the cross-dataset setting. The detailed results for these generalization tests, along with additional performance analysis, are available in the Supplementary Material (Section~\ref{sec:generalize}).

\subsection{Ablation Studies}
We individually evaluated each component of our volume estimation system, analyzing the impact of image features, text features, and the encoder backbone, and text prompts. 

\subsubsection{Multi-Modal Feature Combination} \label{sec:feature_combination}
\begin{table}[h]
    \centering
    \footnotesize
    \begin{tabular}{c| c c}
    \toprule
         \textbf{Module} & \textbf{MAE (mL)} $\downarrow$ & \textbf{MAPE (\%)} $\downarrow$\\
         \midrule
         Stereo Images Only & 77.33 & 38.35 \\
         Text Only & 188.50 & 100.00 \\
         \textbf{Stereo + Text (Ours)} & \textbf{47.95} & \textbf{22.54}\\ 
         \bottomrule
    \end{tabular}
    \caption{\textbf{Importance of Our Multi-Modal Approach.} Using only the text features doesn't provide the model with much information but the combination of stereo features with text outperforms each individual component.}
    \label{tab:multi-modal}
\end{table}

The Text Only approach essentially contains only the class conditioned prior information. The images only uses no class conditioned prior. The performance difference between these is indicative of how our method doesn't overly rely on the class conditioned mean and instead actually uses the visual features more for inference. 
The VLM-inspired projection of image and text features to simulate the human vision and cognition system is the driving force behind the outstanding performance of our model. 
Table~\ref{tab:multi-modal} shows that our model can achieve a lower MAPE when our text-guided stereo image estimator is used. 

\subsubsection{Number of Images}
\begin{table}[h!]
    \centering
    \resizebox{\linewidth}{!}{
    \begin{tabular}{c| c c c}
    \toprule
         \textbf{\# of Images} & \textbf{MAE (mL)} $\downarrow$ & \textbf{MAPE (\%)} $\downarrow$ & \textbf{GFLOPs} $\downarrow$\\
         \midrule
         1 image & 71.25 & 35.11 & 0.12\\
         2 images & 58.72 & 31.49 & 0.17 \\
         5 images & 62.10 & 31.62 & 0.32\\ 
         10 images & 63.31 & 30.91 & 0.57\\ 
         20 images & 53.57 & 29.68 & 1.07\\ 
         50 images & 51.54 & 26.74 & 2.58\\
         \bottomrule
    \end{tabular}}
    \caption{\textbf{Tradeoff between number of images and performance.} Our stereo based approach meant to leverage current stereo image capture devices achieves an incredible tradeoff between performance and computation.}
    \label{tab:num_images}
\end{table}

While our method emulates the stereo vision system of humans to leverage the increasing amounts of egocentric data provided by wearable devices (stereo data), our method can also work with multiple images. We show in Table~\ref{tab:num_images} that increasing the number of images leads to a slight but insignificant improvement in performance at the cost of more computations quantified as GFLOPs. The stereo image setting provides a good balance between performance and computational cost. Furthermore, it allows us to stay on the core premise of the method to use the stereo vision system to leverage new stereo data from wearable devices and smartphones. 

\begin{table*}[ht!]
\centering
\resizebox{\linewidth}{!}{
    \begin{tabular}{c|c c c c c}
    \toprule
    \textbf{Method} & Volume MAE (mL) $\downarrow$ & Energy MAE (kCal) $\downarrow$ &  Protein MAE (g) $\downarrow$ & Carb MAE (g) $\downarrow$ & Fat MAE (g) $\downarrow$ \\
    \midrule
    Baseline & 153.89 & 245.88 & 12.01 & 20.45 & 13.29 \\
    Category Mean & 120.17 & 138.09 & 5.86 & 16.33 & 5.69 \\
    3D Assisted~\cite{Vinod20243DFoodPortion} & 140.01 & 146.76 & 7.65 & 14.26 & 6.75 \\
    GPT-5 (w/context)~\cite{wei2022chain} & 87.69 & 115.42 & 7.02 & 9.81 & 5.38 \\
    RGB Only~\cite{thames2021nutrition5k} & 151.98 & 192.90 & 10.43 & 17.44 & 9.24 \\
    Stereo Recon.~\cite{Dehais2017TwoViewFoodReconstruction} & 137.14 & 193.95 & 10.84 & 16.16 & 9.78 \\
    MFP3D~\cite{ma2024mfp3d} & 62.60 & 77.98 & - & - & - \\
    \textbf{Ours} & \textbf{47.95} & \textbf{60.99} & \textbf{2.55} & \textbf{6.78} & \textbf{2.70} \\
    \bottomrule
    \end{tabular}}
    \caption{\textbf{Dietary Assessment.} Our method accurately predicts the nutritional content of the foods in the image based on our superior volume estimation by achieving the lowest MAE as compared to other established dietary assessment methods. For MFP3D~\cite{ma2024mfp3d}, only the values reported in the paper are shown.}
    \vspace{-0.2cm}
\end{table*}

\subsubsection{Vision Feature Extraction Backbone}
\begin{table}[h]
    \centering
    \small
    \begin{tabular}{l|c c}
    \toprule
       \textbf{Backbone}  & \textbf{MAE (mL)}  $\downarrow$ &\textbf{ MAPE (\%)} $\downarrow$ \\
       \midrule
        DeiT Small & 86.88 & 39.13 \\
        DeiT Base & 67.19 & 34.07 \\
        % ResNet50  & 77.22 & 53.5 \\
        % ViT B/16 & 98.84 & 59.46 \\
        ViT B/32 & 67.04 & 44.56 \\
        ViT L/14 & 66.95 & 30.75 \\
        ResNet50 - CLIP & 89.70 & 47.84 \\
        ResNet101 - CLIP & 91.47 & 46.97 \\
        % ViT B/16 - CLIP & 83.65 & 62.96 \\
        ViT B/32 - CLIP & 70.62 & 39.85 \\
        ViT L/14 - CLIP & 84.44 & 42.43 \\
        \textbf{ViT L/14@336px - CLIP} & \textbf{60.22} & \textbf{31.72} \\
        \bottomrule
    \end{tabular}
    \caption{\textbf{Comparison of feature extractors.} Transformer-based models, especially the CLIP ViT-L/14@336px variant, provide the most discriminative features for volume estimation.}
    \label{tab:backbone_analysis}
\end{table}

The vision features extracted in our model play a crucial role in the volume estimation capabilities of our method. Here, we evaluate different backbones for feature extraction to understand the most suitable extraction backbone. Table~\ref{tab:backbone_analysis} shows that the large Vision transformer models trained using CLIP~\cite{radford2021learning} outperformed other backbones, demonstrating the richness of the features extracted from these models in accurate volume estimation.

\subsection{Text Prompt Analysis}
\begin{table}[h!]
    \centering
    \begin{tabular}{c| c c}
    \toprule
         & \textbf{MAE (mL)} $\downarrow$ & \textbf{MAPE (\%)} $\downarrow$\\
         \midrule
         Text prompt 1 & 59.56 & 34.98 \\
         Text prompt 2 & 61.98 & 34.05 \\
         Text prompt 3 & 65.81 & 35.26\\ 
         Text prompt 4 & 64.27 & 35.54\\ \hline
         Text prompt 5 (\textit{ours)} & 61.85 & 30.88 \\
         \bottomrule
    \end{tabular}
    \caption{Performance of different text prompts. The results demonstrate how variations in prompt formulation affect the model’s predictive performance.}
    \label{tab:text_analysis}
\end{table}

We analyze how variations in the text prompts bring a difference in the final estimated volume. While there are infinitely different possibilities, we used Anthropic's Claude Prompt Generator to refine 4 other versions of our final used prompt. 

Below are the different versions of the prompt generated 
\begin{enumerate}
    \item \textbf{Prompt 1 }- Detected object: \{\ predicted class name \}\, estimated volume: \{\ mean volume \}\ mL
    \item \textbf{Prompt 2} - Object identified as  \{\ predicted class name \}\, with volume approximately \{\ mean volume \}\ mL
    \item \textbf{Prompt 3} - Classification: \{\ predicted class name \}\, $\vert$ Volume estimate: \{\ mean volume \}\ mL
    \item \textbf{Prompt 4} - This appears to be a \{\ predicted class name \}\, measuring roughly \{\ mean volume \}\ mL in volume
    \item \textbf{Prompt 5 (\textit{ours})} - The object is \{predicted class name\} and the approximate volume is \{mean volume\} mL
\end{enumerate}

The variation in MAE (mL) and MAPE (\%) values in Table~\ref{tab:text_analysis} across different text prompts can be attributed to how effectively each prompt formulation aligns the text embedding with the visual features in the shared multimodal space. Even though the underlying semantic information (predicted class and mean volume) is the same, the phrasing of the prompt influences how the language model represents this information in its embedding space.
Prompts that are more concise, structured, and aligned with the model’s pretraining distribution (e.g., simple declarative statements) tend to produce embeddings that cluster more tightly with the corresponding image features, leading to improved regression performance. For instance, our final prompt (Prompt 5) yields the lowest MAPE (30.88\%), suggesting that its phrasing better captures proportional differences in volume estimation, even though its MAE is slightly higher than Prompt 1. This indicates that while Prompt 1 produces a lower absolute error in some cases, Prompt 5 generalizes more consistently across varying object sizes.

% Conversely, prompts that include additional narrative framing (e.g., Prompt 4) introduce linguistic tokens that may dilute the semantic weight of the class and volume terms. This results in embeddings that are less discriminative with respect to volume, thereby increasing both MAE and MAPE.
% In summary, these differences highlight the sensitivity of multimodal models to prompt design. Even minor linguistic variations can shift the embedding space, which directly impacts downstream regression accuracy. Careful prompt engineering thus remains essential to fully exploit text priors in multimodal tasks.

\section{Practical Application}
The MetaFood3D dataset~\cite{chen2024metafood3d} contains specific food codes for each item in the dataset that is obtained from the United States Department of Agriculture (USDA) called the FNDDS food code~\cite{montville2013usda}. This maps each food code to its corresponding nutritional content. However, the values are normalized. We show that our volume estimates can be used to scale the values in the FNDDS dataset and estimate the nutritional content of the food in the images. We compare our estimates with the ground-truth values provided in the dataset. We use the volume estimates from the other methods as well to show a comparison of the potential of our proposed method to be used in solving real-world problems.

We also include other existing dietary assessment methods such MFP3D~\cite{ma2024mfp3d} from the MetaFood3D~\cite{chen2024metafood3d} paper. Our volume estimation method allows accurate image-based dietary assessment and can serve as a strong benchmark for accurate stereo dietary assessment.
\section{Conclusion and Discussion}
\label{sec:conclusion}

In this paper, we introduced a new, multi-modal framework for volume estimation that is inspired by human cognition. Our approach fuses rich geometric cues from stereo vision with semantic context from textual priors. By designing a model architecture with a Vision-Language Model (VLM)-inspired projection layer, we created a method tailored for this challenging geometric regression task. Our experiments show that this method significantly outperforms existing monocular, stereo, and multi-view techniques on public datasets and validates our core hypothesis that combining geometric and semantic data leads to more accurate predictions.

Despite its strong performance, key avenues for future work remain. The primary limitations include the model's computational complexity, which prevents real-time inference on wearable hardware, and the lack of ``in-the-wild'' training data from unconstrained scenarios. Future research will focus on model optimization for on-device deployment and the collection of more diverse datasets to improve real-world generalizability. Ultimately, our work highlights the promising potential of applying VLM-inspired architectures to complex 3D reasoning and  regression tasks.
{
    \small
    \bibliographystyle{ieeenat_fullname}
    \bibliography{main}
}

\clearpage
\setcounter{page}{1}
\maketitlesupplementary

\section{GPT-5 Experiment Prompts}
\label{sec:gpt-5}

For our experiments with GPT-5, the prompts are designed to ensure clarity in how the model is instructed and what data it receives. We used two different prompting structures depending on whether monocular (single image) or stereo (two images) were used for volume estimation.

\paragraph{Single Image Prompt}
The single-image prompt asks the model to simply estimate the volume and ensure a numerical output. Listing \ref{lst:mono_prompt} showcases a ``context-free" version, whereas Listing \ref{lst:mono_prompt_context} is the ``context-aware" alternative.

\begin{lstlisting}[style=jsonbw, caption={Single Image Volume Estimation Prompt (no context)}, label={lst:mono_prompt}]
"Answer with ONLY a single floating-point number (milliliters). No units, no extra text.
Estimate the object's volume in milliliters from the image.
Return ONLY a single floating-point number (milliliters), no units, no words, no punctuation, no JSON, no code fences."
\end{lstlisting}

\begin{lstlisting}[style=jsonbw, caption={Single Image Volume Estimation Prompt (with context)}, label={lst:mono_prompt_context}]
"Answer with ONLY a single floating-point number (milliliters). No units, no extra text.
Given this is an image of {context_text}, estimate its volume in milliliters.
Return ONLY a single floating-point number (milliliters), no units, no words, no punctuation, no JSON, no code fences."
\end{lstlisting}

\paragraph{Stereo Image Prompt}

For the two-view task, a more detailed and structured prompt is passed to the model to rely on stereo cues and return a single numeric estimate. The prompt, in detail, is found in Listing \ref{lst:stereo_prompt}

\begin{lstlisting}[style=jsonbw, caption={Stereo Image Volume Estimation Prompt}, label={lst:stereo_prompt}]
"You are given TWO images of the SAME object, captured from different viewpoints.
Use both images jointly (stereo cues, parallax, shape consistency) to estimate the object's volume in milliliters.
Assume similar scale and camera distance; modest viewpoint change is present.
RESPONSE FORMAT (STRICT JSON, one object, no code fences, no extra text):
{\n  \"volume_ml\": <float>,\n  \"explanation\": \"<2-4 concise sentences on the visual cues you used>\"\n}
Rules:
- Return ONLY the JSON object above (no markdown, no reasoning sections, no additional keys).
- \"volume_ml\" MUST be a single floating-point number (no units, no commas).
Return ONLY the final JSON object; do not include chain-of-thought or extra text."
\end{lstlisting}

\section{Generalizability} \label{sec:generalize}
To evaluate our model's ability to generalize to unseen object categories, we performed experiments on the MetaFood3D dataset~\cite{chen2024metafood3d}. We used a random train-test split, resulting in 415 training and 104 testing samples, which ensures that the test set contains categories absent from training. Given the dataset's limited size, strong generalization performance is not expected. As shown in Table~\ref{tab:random_split_results}, we compare our method against the trainable ``RGB Only'' baseline under these challenging conditions.

\begin{table}[h!]
    \centering
    \begin{tabular}{c|c c}
    \toprule
         \textbf{Method} &  \textbf{MAE (mL)} & \textbf{MAPE (\%)} \\
         \midrule
         Baseline        &   150.02           &     553.96         \\
         RGB Only        &   221.07           &     299.52          \\
         \textbf{Ours}   &   \textbf{40.50}   &     \textbf{22.38} \\ 
         \bottomrule
    \end{tabular}
    \caption{\textbf{Random Split Generalization.} The random split into training and testing splits ensures that there are categories and items that the model has never ``seen'' and yet the performance is significantly better than the RGB estimation method which was also trained with the same training and testing split.}
    \label{tab:random_split_results}
\end{table}

To further assess the model's zero-shot generalization capabilities, we trained our method exclusively on the OmniObject3D dataset~\cite{wu2023omniobject3d} and evaluated it directly on the MetaFood3D dataset~\cite{chen2024metafood3d}. This cross-dataset evaluation introduces a significant domain shift, as the model must contend with numerous out-of-distribution (OOD) images and categories. As presented in Table~\ref{tab:general_results}, while performance is understandably limited by this domain gap, our method still surpasses the "RGB Only" baseline. This result indicates a greater understanding of object sizes and context cues, though achieving true zero-shot performance will necessitate training on larger and more varied datasets.

\begin{figure}
    \centering
    \includegraphics[width=0.8\linewidth]{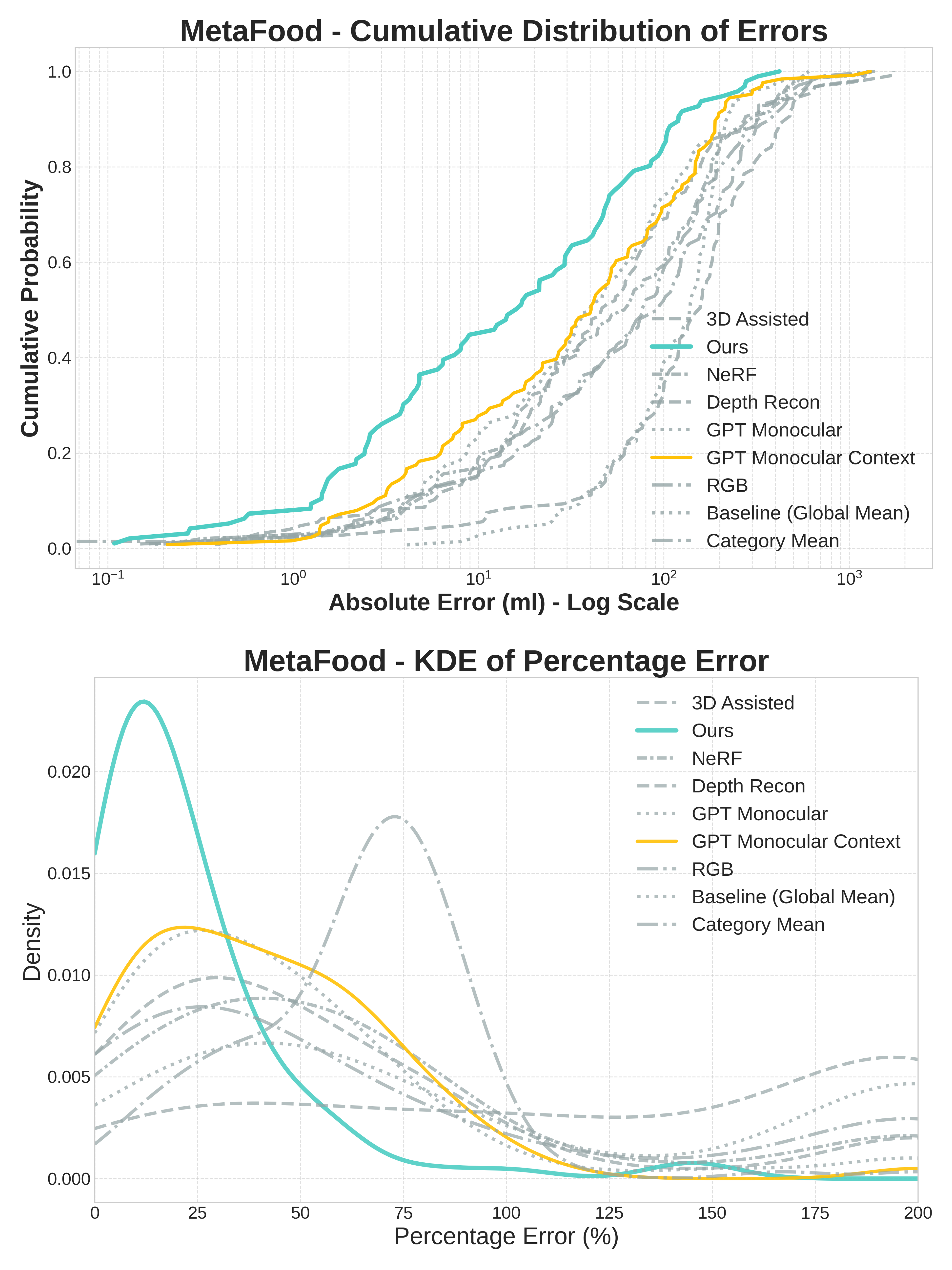}
    \caption{\textbf{Error Distribution of Volume Estimation Methods.} We highlight our method (blue) against the second-best performing method, GPT-5 with context (orange), with regards to MAPE. \textbf{(Top)} The Cumulative Distribution Function (CDF) of absolute errors on a log scale where a shift or curve to the upper left indicates better performance. \textbf{(Bottom)} The Kernel Density Estimation (KDE) of percentage errors where a tall peak near zero with significant delay indicates better performance.}
    \label{fig:big_metafood_2}
\end{figure}

\begin{table}[h!]
    \centering
    \begin{tabular}{c|c c}
    \toprule
         \textbf{Method} &  \textbf{MAE (mL)} & \textbf{MAPE (\%)}\\
         \midrule
         Baseline        &   165.75           &  836.50           \\
         RGB Only        &   174.57           &  212.96           \\
         GPT-5           & 92.38   & 198.83 \\
         \textbf{Ours}            &   \textbf{96.08}           &   \textbf{178.86}           \\  
         \bottomrule
    \end{tabular}
    \caption{\textbf{Cross Dataset Validation.} The model tested on ``out-of-distribution'' images still outperforms the baseline and the RGB Only method which was trained and tested on the same data as our method.}
    \label{tab:general_results}
\end{table}

\section{Additional Error Visualizations} \label{sec:additional_visuals}

\begin{figure}[htbp]
    \centering
    \includegraphics[width=0.8\linewidth]{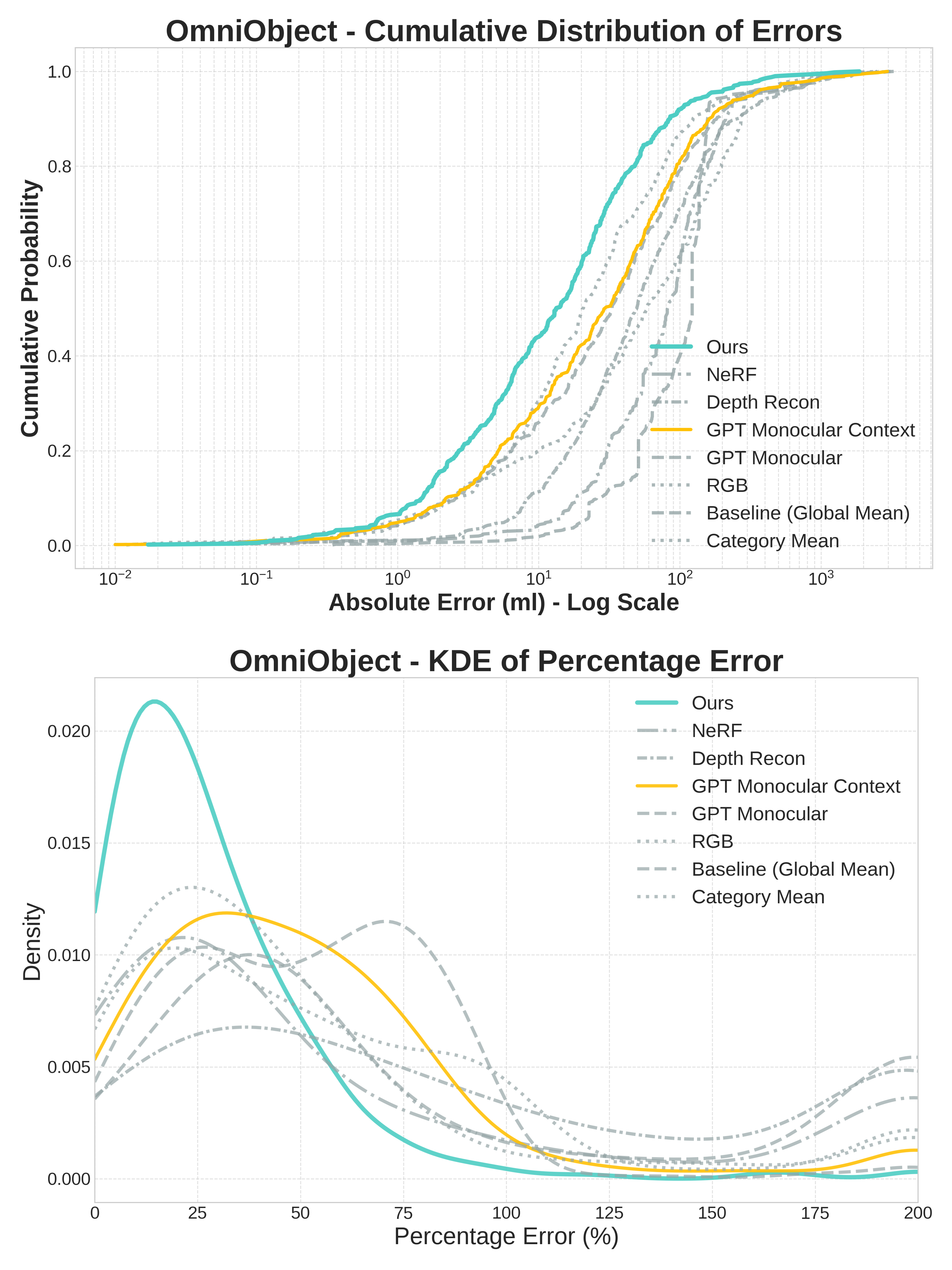}
    \caption{\textbf{Error Distribution of Volume Estimation Methods.} We highlight our method (orange) against the second-best performing method, GPT-5 with context (blue), with regards to MAPE. \textbf{(Top)} The Cumulative Distribution Function (CDF) of absolute errors on a log scale where a shift or curve to the upper left indicates better performance. \textbf{(Bottom)} The Kernel Density Estimation (KDE) of percentage errors where a tall peak near zero with significant delay indicates better performance.}
    \label{fig:big_omniobject}
\end{figure}

In Figure~\ref{fig:big_omniobject}, we perform the same error analysis for OmniObject as we did in Figure~\ref{fig:big_metafood_2} (From main paper). We similarly observe that our method outperforms all other monocular methods, having more effective absolute error and absolute percentage error distributions.

\begin{figure}[htbp]
    \centering
    \includegraphics[width=\linewidth]{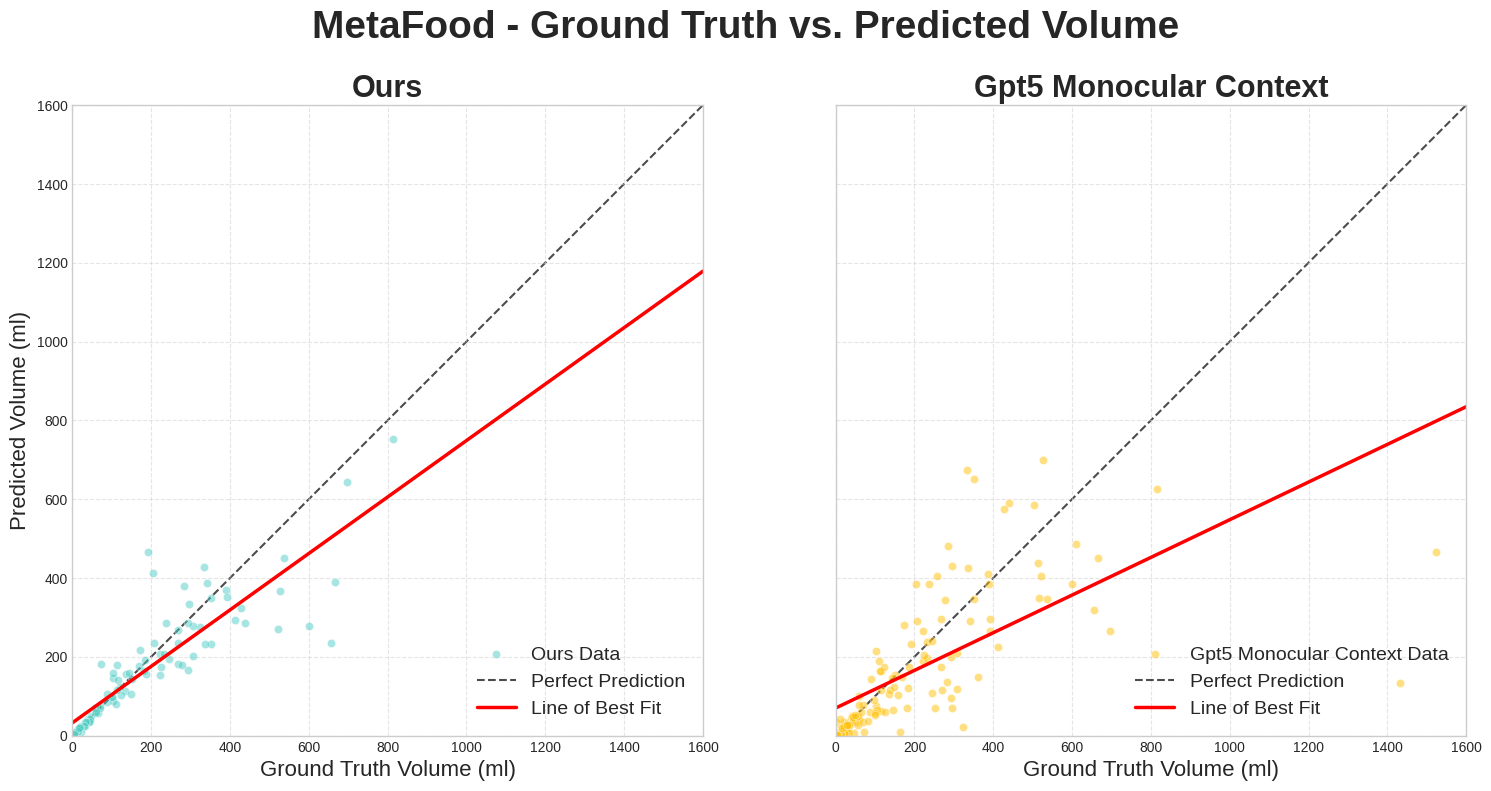}
    \caption{Predicted volumes plotted against ground truth volumes for MetaFood. We plot our method against the second-best performing monocular method, GPT-5 with context, and overlay a line of best fit to help visualize correlation with a ``perfect estimation" line.}
    \label{fig:lobf-MF}
\end{figure}

\begin{figure}
    \centering
    \includegraphics[width=\linewidth]{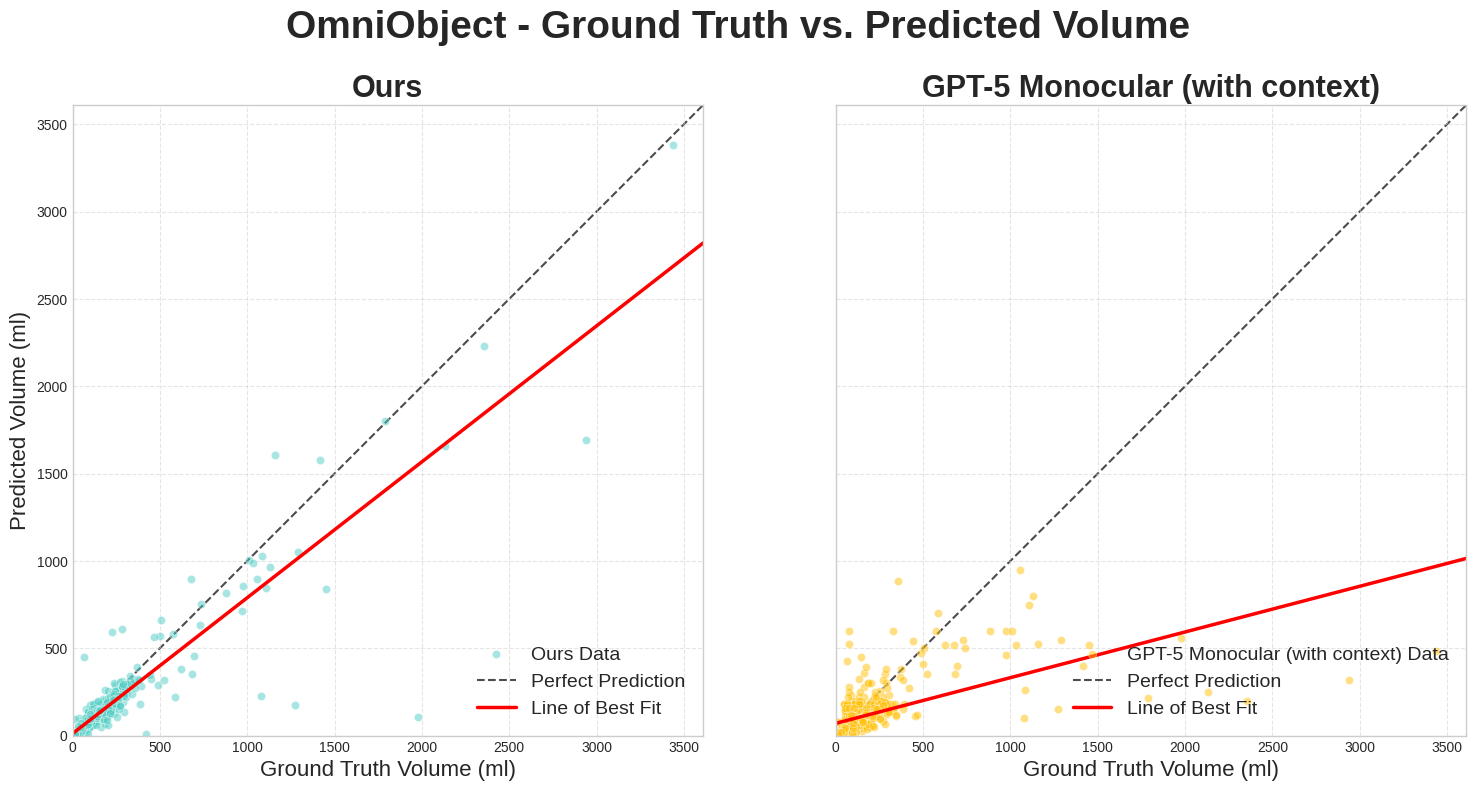}
    \caption{Predicted volumes plotted against ground truth volumes for OmniObject. We plot our method against the second-best performing monocular method, GPT-5 with context, and overlay a line of best fit to help visualize correlation with a ``perfect estimation" line.}
    \label{fig:lobf-OO}
\end{figure}

Figures~\ref{fig:lobf-MF} and \ref{fig:lobf-OO} plot the estimated volumes against the ground truth volumes for MetaFood and OmniObject, respectively. Similar to our error analysis figures, we also compare our method against the second-best performing method in GPT-5 with context. A ``line of best fit" is overlaid these scatter plots to help visualize the correlation between the estimations and a ``perfect estimation" line. We observe how, for both datasets, our method's line of best fit is more closely aligned to the perfect estimation line.

\end{document}